\documentclass[10pt,twocolumn,letterpaper]{article}

\usepackage[pagenumbers]{cvpr}

\usepackage{amsmath,amssymb}
\usepackage{booktabs}
\usepackage{graphicx}
\usepackage{xspace}
\usepackage{multirow}
\usepackage{algorithm}
\usepackage{algpseudocode}
\usepackage{comment}
\usepackage{makecell}
\usepackage{pifont}
\newcommand{\xmark}{\textcolor{red}{\ding{55}}}
\newcommand{\cmark}{\textcolor{ForestGreen}{\ding{51}}}

\newcommand{\brainsg}{BrainSG\xspace}

\newcommand{\brainsgbench}{BrainSG-Bench\xspace}
\DeclareMathOperator*{\argmaxop}{arg\,max}

\definecolor{cvprblue}{rgb}{0.21,0.49,0.74}
\usepackage[pagebackref,breaklinks,colorlinks,allcolors=cvprblue]{hyperref}

\def\confName{CVPR}
\def\confYear{2027}

\title{BrainSG: Beyond Object to Relation Understanding via  \\ Multimodal Vision-Brain Scene Graph Generation}

\author{
Hoang-Quan Nguyen$^1$, Quyen-Nang Bui$^1$, Suhyun Kim$^2$, Hojin Jang$^2$, Khoa Luu$^1$\\
$^1$Department of Electrical Engineering and Computer Science, University of Arkansas, AR, USA\\
$^2$Department of Brain and Cognitive Engineering, Korea University, Seoul, South Korea\\
\tt\small{\{hn016, nbui, khoaluu\}@uark.edu, \{ksuuueh0224, hojin4671\}@korea.ac.kr} \\
\tt\small\url{https://uark-cviu.github.io/projects/brainsg} 
}

\begin{document}
\maketitle
\begin{abstract}
Decoding scene structure from brain activity requires recovering not only which objects are present, but also how they relate. Existing brain-decoding benchmarks mainly evaluate reconstructed images or videos, predicted object categories, captions, or global image-text embeddings.
These evaluations can show that a decoder recovers the objects and general meaning of a scene, but they do not explicitly evaluate whether two objects are connected by the correct predicate, whether the relation is reversed, or whether its roles are assigned to the correct object instances.
To address this gap, we first introduce \brainsgbench, a new structured vision-brain scene-graph dataset and evaluation benchmark.
It provides two forms of structured relational supervision that prior brain-decoding benchmarks lack: human-audited directed scene graphs and neuroscience-expert annotations of hypothesized cognitive functions and regional involvement.
\brainsgbench supports three complementary tasks: graph retrieval, relation decoding with known objects, and complete scene-graph generation.
In addition, to address limited paired supervision and variable signal reliability, we propose \brainsg, a brain-to-graph model with noise-adaptive prior attention.
The model represents fMRI signals as ROI tokens and estimates ROI-token reliability directly from the input brain signals during inference, adapting predicate-conditioned attention guided by a fixed predicate-to-ROI prior.
On \brainsgbench, \brainsg achieves 45.6\% Recall@5 for graph retrieval and triplet F1 scores of 57.4\% for relation decoding with known objects and 31.1\% for scene-graph generation.
Together, \brainsgbench and \brainsg establish a new framework for studying how structured supervision of brain signals can support relational decoding.\footnote{Code, model, and dataset will be published to the research community.}
\end{abstract}

\begin{figure}
    \centering
    \includegraphics[width=\linewidth]{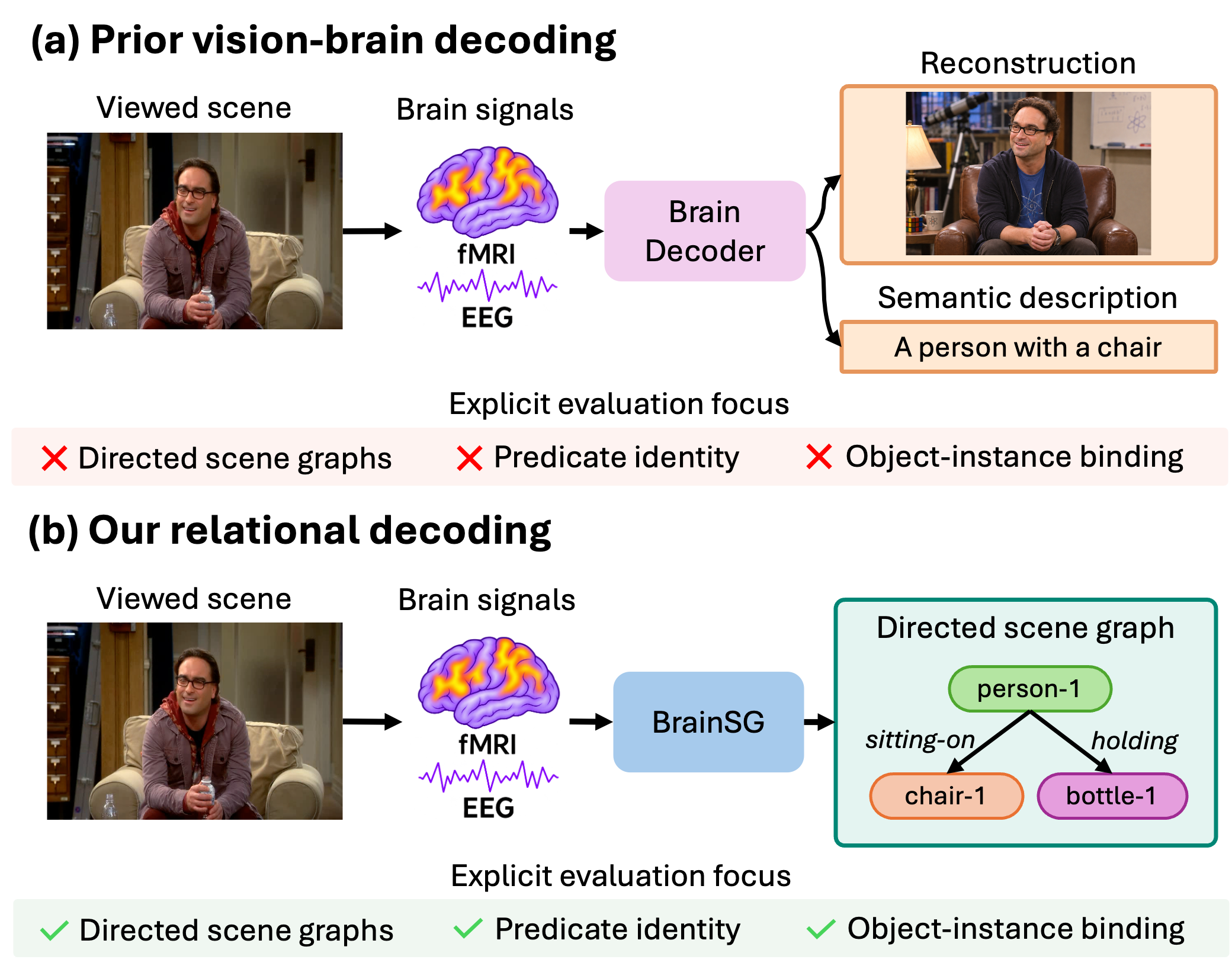}
    \vspace{-6mm}
    \caption{
    \textbf{From vision-brain content decoding to relational decoding.} 
    (a) Prior methods: given fMRI and EEG, they identify objects or generate semantic descriptions. 
    (b) Our BrainSG predicts directed scene graphs, explicitly evaluating predicates, relation direction, and object-instance assignment.
    }
    \label{fig:motivation}
    \vspace{-6mm}
\end{figure}

\vspace{-3mm}
\section{Introduction}
\label{sec:intro}

Understanding a scene involves recognizing its objects and interpreting how they interact.
In Figure~\ref{fig:motivation}, the description identifies a person and a chair, while the directed scene graph specifies that the person sits on the chair and holds a bottle.
Recovering these relationships requires identifying the predicate, its direction, and the objects assigned to each role.
Brain-decoding research predicts semantic and perceptual features, matches global embeddings, and reconstructs images or videos from neural measurements~\cite{takagi2023highresolution,ozcelik2023braindiffuser,gao2026cinebrain}.
However, these evaluations do not isolate relational distinctions between the objects. 
In this work, we study whether \textbf{\textit{functional magnetic resonance imaging (fMRI) and electroencephalogram (EEG) support recovery of these structured relationships}} in a viewed event.

Scene graphs provide an explicit prediction target by representing objects as nodes and directed relations as labeled edges~\cite{xu2017scenegraph,li2022sgtr}.
Image and video scene-graph methods infer this structure directly from pixels~\cite{ji2020actiongenome,yang2023pvsg}.
A brain decoder instead receives only the corresponding fMRI and EEG responses. This setting asks whether neural measurements can distinguish graphs that contain the same objects but differ in predicate, direction, or subject-object assignment. Such controlled graph comparisons complement scene reconstruction: reconstruction captures overall visual and semantic content, whereas graph decoding isolates the relationships among particular objects.
This distinction is important because a pretrained generator can produce plausible interactions from its generative priors even when the neural measurement supports only coarse object-level semantics.
Query-conditioned methods decode interacting concepts and predicate-argument structure from fMRI, but the query specifies which concepts or relations to examine~\cite{wang2026neurona}.
We focus on the complementary setting of query-free recovery or ranking of a directed scene graph.

We first introduce \textbf{\textit{\brainsgbench}}, a \textit{novel structured vision brain scene-graph dataset and evaluation benchmark for brain-to-graph prediction}.
It builds on synchronized fMRI and EEG recordings collected by Gao \etal during audiovisual observation~\cite{gao2026cinebrain}.
The source dataset was designed for multimodal brain decoding and video reconstruction, leaving relation-level annotation and evaluation outside its original scope.
From this foundation, \brainsgbench constructs temporally aligned brain-to-graph examples with human-audited directed graphs, instance-level subject-object bindings, relation evidence labels, and neuroscience-expert functional annotations.
The dataset also provides controlled relational alternatives, fixed splits, and standardized evaluators.
For each four-second interval, the decoder receives the aligned fMRI response, EEG response, or both, without access to the source video at test time.
We assess this output at three levels: ranking a complete graph among controlled alternatives, decoding relations when the objects are given, and generating the complete graph from neural input.
Relations are annotated from visual, auditory, or contextual evidence.
Expert annotations describe hypothesized cognitive operations, temporal demands, and functional-network or region-of-interest (ROI) involvement.
Section~\ref{sec:benchmark} details our proposed data, annotations, temporal alignment, and evaluation protocol.

This work also proposes \textbf{\textit{\brainsg}}, a \textit{new brain-to-graph model with noise-adaptive prior attention}. 
To achieve this goal, the decoder must identify which regional and temporal measurements support each candidate relation, even though fMRI and EEG signals vary in reliability across observations.
Multimodal fusion methods use cross-attention to model dependencies between modalities~\cite{truong2025mango,truong2025directed}.
Because different predicates may draw on different neural patterns, a predicate-to-ROI prior guides relation queries toward regions plausibly associated with each predicate.
Prior-informed attention naturally introduces this structure into learned representations.
Bayesian Attention Modules treat attention weights as stochastic latent variables~\cite{fan2020bayesian}, Graphormer incorporates graph structure through additive attention biases~\cite{ying2021transformers}, and Grounding-based Attention Priors learn concept-to-region correspondences~\cite{le2023guiding}.
However, these methods do not address brain scene-graph decoding with input-dependent signal reliability.
With limited paired supervision, the decoder must learn which brain regions support each relation while accounting for signal-quality variation across inputs. Thus, this work presents \brainsg, a new brain scene-graph decoder, to balance prior structure with evidence from measured brain signals without requiring externally supplied quality scores.

\vspace{1mm}
\noindent
\textbf{Contributions of this Work:} This work presents \textit{one of the first studies on vision-brain scene-graph generation}.
First, we introduce the new problem of brain-to-graph decoding, which recovers directed scene graphs from fMRI and EEG signals to explicitly model predicates, relation direction, and instance-level subject-object bindings.
Second, we present \brainsgbench, which, to the best of our knowledge, is \textit{one of the first datasets for scene-graph decoding} from synchronized fMRI and EEG recordings.
It pairs these recordings with manually \textbf{\textit{human-audited directed graphs, relation evidence labels, and functional annotations by neuroscience experts}}.
The benchmark uses controlled graph alternatives to assess relational recovery on held-out stimuli across neural modalities.
Third, to account for uncertain predicate-ROI correspondence and variable brain-signal reliability, we propose a novel \brainsg approach with noise-adaptive prior attention.
During inference, the mechanism estimates ROI-token reliability directly from the input brain signals and uses these estimates to adapt attention guided by a predicate-to-ROI prior in the model, supporting the decoding of entities and directed relations. Finally, this work provides a standard framework, reproducible models, baseline results, and source code to support future exploration of this problem.

\section{Related Work}
\label{sec:related}

\noindent\textbf{Vision-brain decoding.}
Vision-brain decoding studies recovered perceived visual stimuli from neural activity despite scarce paired data, noisy signals, and variation across participants.
Early work reconstructed images by matching hierarchical network features decoded from fMRI~\cite{shen2019deep}. Later methods mapped fMRI into diffusion latents on natural-image datasets~\cite{takagi2023highresolution,ozcelik2023braindiffuser,chang2019bold5000,allen2022nsd}. Contrastive alignment supported stimulus retrieval from fMRI, EEG, and MEG~\cite{scotti2023mindeye,song2024decoding,benchetrit2024brain}.
Masked pretraining addressed limited paired data~\cite{chen2023mindvis,bai2023dreamdiffusion,chen2023cinematic}, and shared representations supported cross-subject transfer~\cite{wang2024mindbridge,scotti2024mindeye2,beliy2026brain,wang2026zebra}.
Continual learning addressed forgetting across participants in COBRA~\cite{nguyen2026cobra} and bias from shifts across recording sessions in BRAIN~\cite{nguyen2025brain}.
Brainformer used fMRI to guide vision models~\cite{nguyen2025brainformer}, and BRACTIVE aligned visual and neural features to identify subject-specific regions of interest~\cite{nguyen2025bractive}.
Quantum-Brain modeled fMRI voxel dependencies with a quantum-inspired network~\cite{nguyen2024quantum}, and hierarchical quantum control gates encoded local and global fMRI features for object classification~\cite{nguyen2024hierarchical}.
Video decoders extended diffusion over time~\cite{chen2023cinematic}, separated semantic keyframes from motion~\cite{gong2024neuroclips}, or incorporated action and object semantics with retrieved memories~\cite{wei2026bridging}.
EEG and MEG provided millisecond-scale temporal resolution, complementing the spatial specificity of fMRI.
Video datasets extended neural recording to naturalistic events~\cite{lahner2024modeling,gifford2026large}, and CineBrain paired synchronized fMRI and EEG for video reconstruction~\cite{gao2026cinebrain}. Encoding and decoding models also helped characterize the visual cortex~\cite{ozcelik2023braindiffuser,luo2023brain,yang2024braindecodes}.
Typical outputs included pixels, embeddings, captions, or grounded object descriptions~\cite{xia2024umbrae}. Generative priors could yield plausible outputs with ambiguous predicates, directions, or instance bindings. NEURONA decoded interacting concepts with a supplied query~\cite{wang2026neurona}.
\brainsgbench added human-audited directed graphs, controlled alternatives, and expert functional annotations to CineBrain. \brainsg decoded graphs without a supplied query using fMRI, EEG, or both.

\noindent\textbf{Scene graph generation.}
Scene graph generation (SGG) studies modeled directed object relations.
Image SGG built on dense annotations~\cite{krishna2017visualgenome} using message passing, global context, debiasing, and transformer-based graph assembly~\cite{xu2017scenegraph,zellers2018motifs,tang2020unbiased,li2022sgtr}.
Video SGG added temporal structure~\cite{ji2020actiongenome} through tracklets and context~\cite{cong2021sttran,feng2023dsgdetr}, uncertainty-aware and panoptic modeling~\cite{nag2023tempura,yang2023pvsg}, and one-stage prediction~\cite{wang2024oed}.
HIG aggregated interactions hierarchically~\cite{nguyen2024hig}, CYCLO updated interaction histories cyclically~\cite{nguyen2024cyclo}, and HyperGLM used hypergraphs for higher-order interactions and anticipation~\cite{nguyen2025hyperglm}.
\brainsg predicted directed graphs from aligned neural responses.

\section{The Proposed BrainSG-Bench Dataset}
\label{sec:benchmark}

\begin{figure*}
    \centering
    \includegraphics[width=0.9\linewidth]{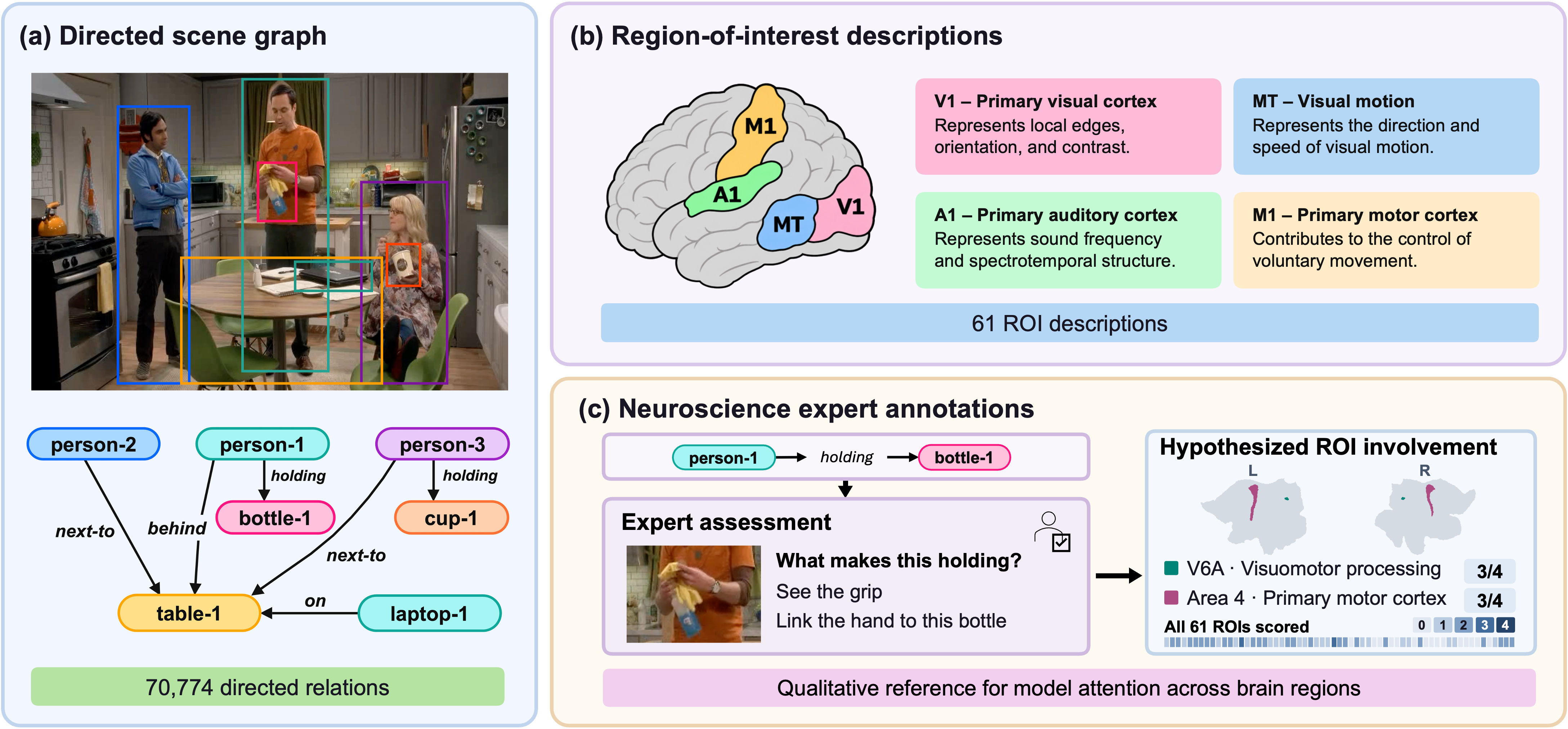}
    \vspace{-2mm}
    \caption{\textbf{Annotation overview of \brainsgbench.} (a) A sample video frame with object instance boxes and directed scene graph. (b) ROI descriptions summarize regional functions. (c) Expert functional hypotheses associate relations with potentially involved brain regions.
    }
    \vspace{-4mm}
    \label{fig:brainsg_bench}
\end{figure*}

\begin{table}[t]
    \centering
    \footnotesize
    \setlength{\tabcolsep}{4pt}
    \caption{
    Comparison of brain datasets and annotation resources.
    Each directed relation is a subject-predicate-object triplet.
    }
    \vspace{-3mm}
    \label{tab:dataset_comparison}
    \resizebox{\linewidth}{!}{\begin{tabular}{@{}lcccc@{}}
    \toprule
    Dataset & \makecell{Neural\\modality} & \makecell{Directed\\relations} & \makecell{ROI\\descriptions} & \makecell{Expert ROI\\hypotheses} \\
    \midrule
    BOLD5000~\cite{chang2019bold5000} & fMRI & \xmark & \xmark & \xmark \\
    NSD~\cite{allen2022nsd} & fMRI & \xmark & \xmark & \xmark \\
    THINGS-fMRI~\cite{hebart2023things} & fMRI & \xmark & \xmark & \xmark \\
    THINGS-EEG2~\cite{gifford2022large} & EEG & \xmark & \xmark & \xmark \\
    CineBrain~\cite{gao2026cinebrain} & fMRI + EEG & \xmark & \xmark & \xmark \\
    \hline
    \textbf{\brainsgbench (Ours)} & fMRI + EEG & \textbf{70,774} & \textbf{61} & \cmark \\
    \bottomrule
    \end{tabular}}
    \vspace{-6mm}
\end{table}

To study relational decoding from brain activity, a dataset pairing neural responses with objects and their directed relations is essential.
In this section, we introduce \brainsgbench, a new structured vision-brain scene-graph dataset and evaluation benchmark.
Specifically, each example pairs synchronized fMRI and EEG responses with a directed scene graph and neuroscience-expert functional hypotheses.
The benchmark tests whether brain responses distinguish the correct scene graph from alternatives containing the same objects but different relationships.
Figure~\ref{fig:brainsg_bench} illustrates an overview of dataset annotation, and Table~\ref{tab:dataset_comparison} compares its scale and annotations with existing brain datasets.

\subsection{Dataset Overview}

\brainsgbench pairs four-second clips with their fMRI and EEG responses, along with directed scene graphs and neuroscience-expert functional hypotheses. Each graph identifies the objects present and their directed subject-predicate-object relations, while the expert annotations describe hypothesized cognitive operations and functional-network or brain region-of-interest (ROI) involvement associated with those relations.
The underlying visual and neural recordings come from CineBrain~\cite{gao2026cinebrain}, while the graph annotations and functional hypotheses contribute the new structured supervision.
The dataset contains 8,100 unique clips, 128,506 object instances, and 70,774 relation triplets.

\begin{figure}
\vspace{-2mm}
    \centering
    \includegraphics[width=\linewidth]{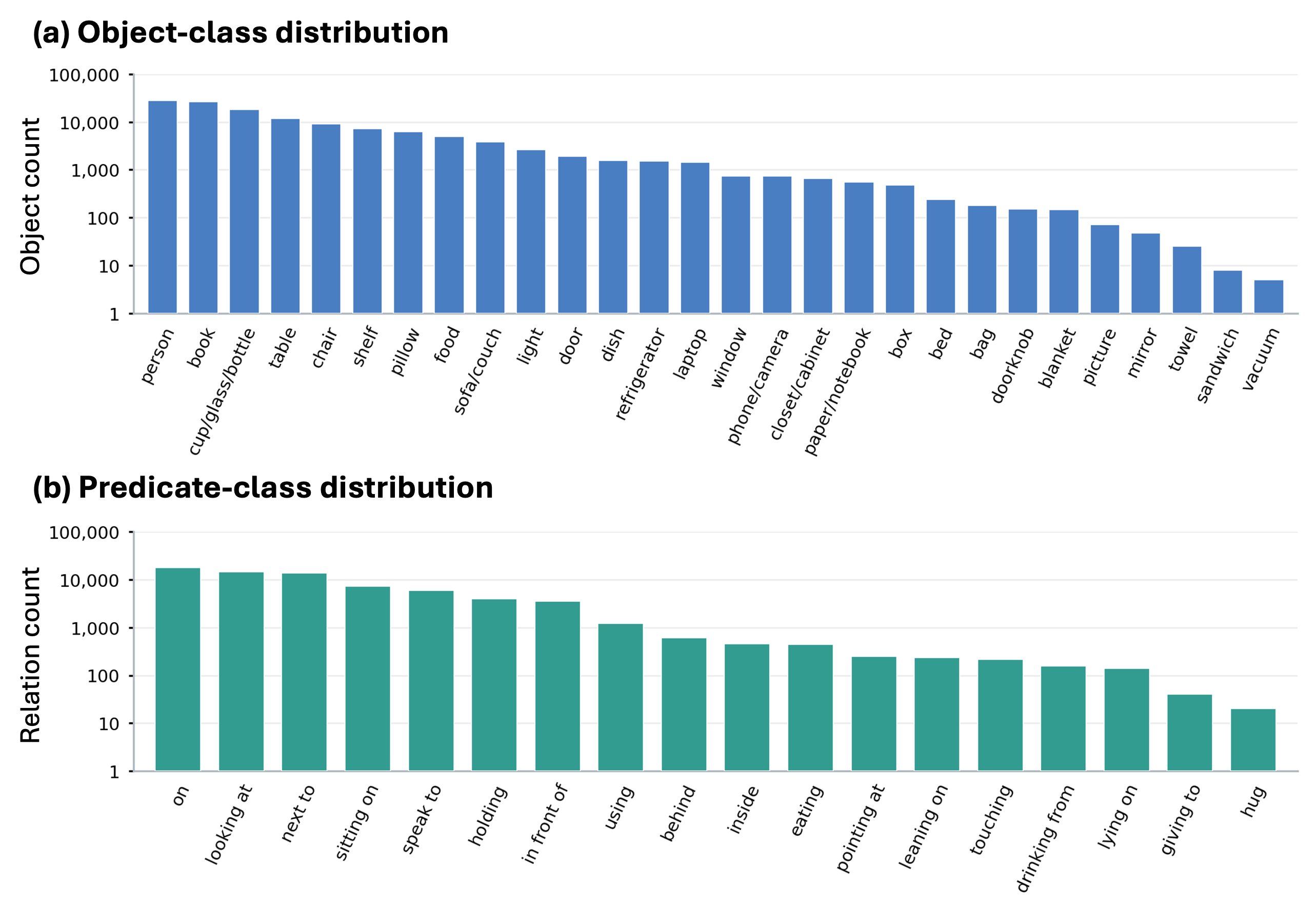}
    \vspace{-8mm}
    \caption{Distribution of object classes (top) and relation predicates (bottom) in \brainsgbench.}
    \label{fig:data_analysis}
    \vspace{-6mm}
\end{figure}

For one participant and one four-second interval, a record is $\mathcal{D}=(X, B^F, B^E, G)$,
where $X$ contains a video, $B^F$ and $B^E$ are the corresponding fMRI and EEG windows, and $G$ is the target graph with associated expert functional annotations.
Neural windows are assigned using a fixed temporal-alignment rule accounting for the delayed fMRI response.
Figure~\ref{fig:data_analysis} shows the distributions of object classes and relation predicates in \brainsgbench.

\subsection{Scene-Graph Annotation}

For each video interval, we annotate a directed scene graph $G=(\mathcal{V},\mathcal{E})$, where nodes represent objects and edges represent subject-predicate-object relations.
Each object has a category and a unique instance ID, allowing objects of the same category to be distinguished.
Each relation specifies its predicate and the IDs of its subject and object, preserving direction and instance assignment.
The relation vocabulary covers spatial, contact, possession, attention, communication, and action relations.

\subsection{Neuroscience-Expert Functional Hypotheses}

Scene graphs describe the objects and their relationships in human brain signals.
To complement these annotations, neuroscience experts formulate hypotheses about the cognitive operations and brain networks or atlas ROIs that may contribute to processing each relation.
For example, \texttt{speaking-to} may involve language processing, social interaction, and distinguishing the speaker from the recipient.
These annotations describe possible functional involvement; they do not establish participant-specific activation or causal localization.
Experts follow a predefined vocabulary and do not inspect held-out neural responses.

\subsection{Evaluation Tasks}

Using the annotated graphs as targets, we define three tasks for evaluating relational decoding.
\textit{Brain-to-graph retrieval} ranks the target among fixed candidates; \textit{relation decoding with known objects} predicts edges, predicates, direction, and binding from the supplied anonymous object inventory; \textit{brain-to-scene-graph generation} predicts both objects and relations.
All tasks use neural inputs without held-out stimulus video or transcripts.
Evaluation metrics are specified in Section~\ref{sec:experiments}.

\noindent\textbf{Brain-to-graph retrieval.}
Given neural responses and a candidate set $\mathcal{C}=\{G^+,G_1^-,\ldots,G_K^-\}$, the model selects the correct graph $G^+$ among $K$ incorrect alternatives.
Where eligible, alternatives contain the same objects but reverse subject-object roles, redirect a relation to another same-category instance, or substitute a compatible predicate.
Graphs from nearby intervals and independently proposed hard negatives provide additional comparisons.

\noindent\textbf{Relation decoding with known objects.}
The model receives neural responses and the correct object categories and instance IDs, then predicts directed relations between the objects.
For each ordered pair, it determines whether a relation exists and identifies its predicate.
This task evaluates relation recovery separately from object recognition, including whether the subject and object are assigned to the correct instances.

\noindent\textbf{Brain-to-scene-graph generation.}
The model predicts object instances, their categories, and directed relations from neural responses without a supplied object inventory or candidate graphs.
The evaluation of this task accounts for missing and incorrect objects and relations.

\section{The Proposed BrainSG Approach}
\label{sec:method}

\begin{figure*}
    \centering
    \includegraphics[width=0.97\linewidth]{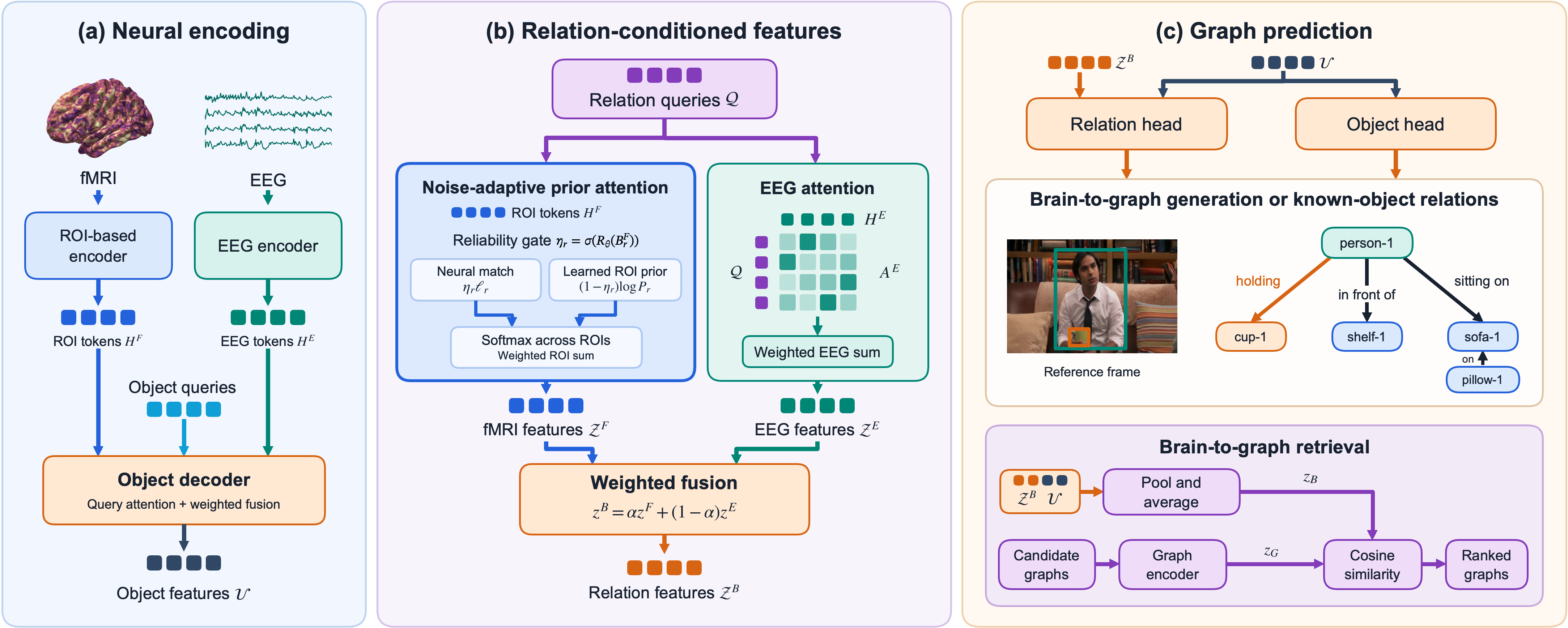}
    \vspace{-2mm}
    \caption{
    \textbf{Overview of the BrainSG framework.} (a) Separate encoders produce fMRI ROI tokens and EEG tokens. (b) Noise-adaptive prior attention balances fMRI evidence with a learned ROI prior according to signal reliability. Weighted fusion combines the resulting relation features with EEG features. (c) BrainSG supports relation decoding, scene-graph generation, and candidate graph retrieval.
    }
    \label{fig:overview_framework}
    \vspace{-4mm}
\end{figure*}

In this section, we introduce \brainsg, a framework for decoding directed scene graphs from fMRI and EEG responses.
As shown in Figure~\ref{fig:overview_framework}, the framework represents fMRI signals as region-of-interest (ROI) tokens and encodes EEG signals with a separate encoder.
Our noise-adaptive prior attention estimates ROI-token reliability to balance measured neural evidence with a fixed functional prior and learns a distribution over regional attention weights.
A weighted fusion combines the fMRI representation with EEG features for graph retrieval, relation decoding with known objects, and complete scene-graph generation.

\subsection{Problem Formulation}

For one participant and one video interval, let $B=(B^F, B^E)$ denote the corresponding fMRI and EEG measurements.
The target is a directed scene graph $G=(\mathcal{V},\mathcal{E})$, where each node represents an object instance with a category and each edge $(v_i,\rho,v_j)$ specifies a predicate $\rho$ from subject $v_i$ to object $v_j$.

We consider the three prediction settings defined in Section~\ref{sec:benchmark}.
For brain-to-graph retrieval, the model scores each candidate in $\mathcal{C}$ and selects
$\widehat{G}=\argmaxop_{G\in\mathcal{C}}s_\theta(B,G)$,
where $s_\theta$ measures compatibility between the neural input and a candidate graph.
For relation decoding with known objects, the model receives object categories and instance IDs, then predicts the existence of relations and predicates for ordered object pairs.
For complete graph generation, it predicts both object instances and directed relations without a supplied inventory or candidate set.

\subsection{ROI-Based fMRI Encoding}

Compared with voxel-level encoding, ROI-based encoding provides a compact representation with explicit anatomical correspondence.
This structure allows the subsequent attention module to learn relation-specific regional preferences and adjust their contributions according to estimated signal reliability.
We therefore construct one fMRI token per ROI.
Each four-second interval is associated with five fMRI samples at a repetition time (TR) of 0.8 seconds.
Let $B^F_r\in\mathbb{R}^{5\times N_r}$ denote these input signals for the $N_r$ voxels in ROI $r$.
Following CineBrain~\cite{gao2026cinebrain}, a shared linear projection maps each voxel's five-sample sequence to a feature vector.
We then average the projected features over valid voxels within each ROI to obtain $g_\theta(B^F_r)$.

We add a learned positional embedding $e_r$ to represent each ROI's anatomical location.
A Transformer models dependencies among the resulting ROI tokens:
\begin{equation}
\footnotesize
    h^{0}_r=g_\theta(B^F_r)+e_r,\quad
    H^F=\operatorname{ROIEncoder}\left(\{h^{0}_r\}_{r=1}^R\right).
\end{equation}
The encoder outputs $R$ contextualized ROI feature vectors of dimension $d$, collected as $H^F\in\mathbb{R}^{R\times d}$ for noise-adaptive prior attention.

\subsection{Noise-Adaptive Prior Attention}

\begin{figure}
    \centering
    \includegraphics[width=0.96\linewidth]{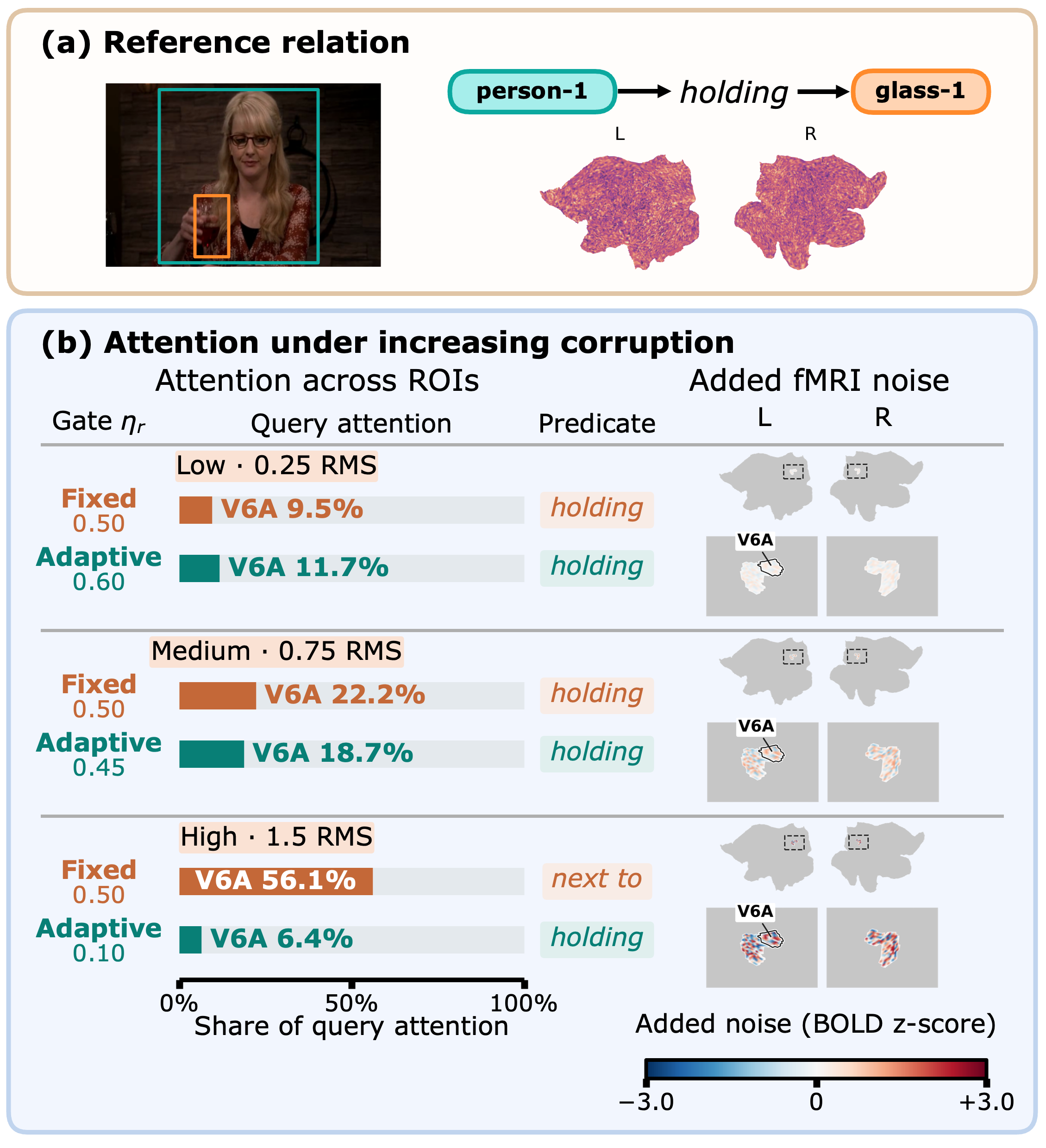}
    \vspace{-3mm}
    \caption{
    \textbf{Noise-adaptive prior attention.} 
    (a) Reference relation and BOLD sample. 
    (b) Added noise is scaled within each affected ROI to a root mean square (RMS) of 0.25, 0.75, or 1.5 in BOLD z-score units. At the highest noise level, the adaptive gate preserves the reference predicate.
    }
    \label{fig:comparison}
    \vspace{-5mm}
\end{figure}

Learning which brain regions support each relation is challenging with limited paired brain-graph data and noisy brain signals.
We therefore learn a relation-ROI prior that captures regional preferences shared across training examples.
Noise-adaptive prior attention adjusts the balance between this prior and measured neural evidence according to estimated signal reliability.
Figure~\ref{fig:comparison} illustrates how the adaptive gate reduces attention to corrupted ROI and preserves the reference predicate under high synthetic noise.

For each predicate $\rho$, trainable logits $A_{\rho,r}\in\mathbb{R}$ define a distribution over ROIs:
\begin{equation}
\footnotesize
    P_{\rho,r}=\frac{\exp(A_{\rho,r})}
    {\sum_{r'}\exp(A_{\rho,r'})}.
\end{equation}
The logits are optimized jointly with the decoder through the graph-decoding objectives.
Unlike input-dependent neural attention, this prior is shared across examples and frozen after training.

For an ordered object pair and a candidate predicate, a relation query attends to the ROI features $h_r\in H^F$.
We omit the object-pair and predicate indices below for clarity.
Let $\ell_r$ be the query's compatibility with ROI $r$, computed by scaled dot-product attention, and let $P_r$ be the learned prior for the candidate predicate.
A learnable function $R_\theta$ maps the fMRI signals $B^F_r$ to a scalar, giving the gate $\eta_r=\operatorname{sigmoid}(R_\theta(B^F_r))$.
We compute attention weights across ROIs as
\begin{equation}
\footnotesize
    a_r=\operatorname{softmax}_r
    \left(\eta_r\ell_r+(1-\eta_r)\log P_r\right).
\end{equation}
Larger gate values favor neural evidence, while smaller values favor the relation-ROI prior.
To supervise this gate, a measurement-quality ranking loss trains $R_\theta$ to score lower-corruption versions of the same ROI signal above higher-corruption ones (see the training objective below).
This supervision establishes a relative ordering under the specified corruptions; reliability under real acquisition noise requires separate validation.
We then aggregate the regional features $z^F=\sum_r a_r W_Vh_r$,
where $W_V$ is a learned projection and $z^F$ is the relation-conditioned fMRI representation.

\subsection{Brain-to-Graph Retrieval and Decoding}

We obtain EEG tokens $H^E$ using the CineBrain encoder architecture~\cite{gao2026cinebrain}.
Let $\mathcal{Q}$ be the set of relation queries, where $q_{ij\rho}$ queries predicate $\rho$ from object $v_i$ to object $v_j$.
The EEG attention matrix $A^E$ has one row per relation query and one column per EEG token. Each row contains the weights used to aggregate EEG token values into the query's EEG feature $z^E$.
For each query, a weighted gate $\alpha\in[0,1]$ combines the fMRI and EEG features as $z^B=\alpha z^F+(1-\alpha)z^E$.
Across relation queries, $\mathcal{Z}^F$, $\mathcal{Z}^E$, and $\mathcal{Z}^B$ collect the fMRI, EEG, and fused features, respectively. 
We write $z^B_{ij\rho}$ for the fused feature of query $q_{ij\rho}$.
The relation head uses these features to predict individual relations.

\noindent\textbf{Object and relation decoding.}
For graph generation, learned object queries attend to the fMRI and EEG features. The fusion module combines their modality representations into the object feature collection $\mathcal{U}=\{u_i\}_{i=1}^{Q}$, where $Q$ is the number of object slots.
Prediction heads map $u_i$ to object categories and existence scores, with bipartite matching during training.
For known-object decoding, embeddings of the supplied categories and instance IDs initialize the object queries instead.
For each ordered pair, predicate queries constructed from $u_i$ and $u_j$ obtain fused features $z^B_{ij\rho}$ through prior attention and multimodal fusion.
A shared prediction head scores each predicate:
\begin{equation}
\footnotesize
    c_{ij\rho}=f_{\mathrm{rel}}
    \left([u_i;u_j;z^B_{ij\rho}]\right),
\end{equation}
where semicolons denote concatenation and $f_{\mathrm{rel}}$ is a learned scalar-output head.

\noindent\textbf{Brain-to-graph retrieval.}
For brain-to-graph retrieval, learned queries extract relation features $\mathcal{Z}^B$ and object features $\mathcal{U}$ from the neural input before scoring any candidate.
We average each collection and give the two means equal weight in the shared brain embedding:
{
\footnotesize
\begin{align}
    z_B&=\tfrac{1}{2}\left(\operatorname{Pool}(\mathcal{Z}^B)
        +\operatorname{Pool}(\mathcal{U})\right),\\
    s_\theta(B,G)&=\operatorname{cos}
    \left(W_Bz_B,W_Gz_G\right).
\end{align}
}
The graph encoder constructs $z_G$ by pooling object and directed-edge features, and $W_B$ and $W_G$ denote learned projection heads.
The highest-scoring graph is retrieved from a candidate set constructed independently of test brain activity.

\subsection{Training Objective}

\noindent\textbf{Graph retrieval.}
For a neural input $B$, target graph $G^+$, and candidate set $\mathcal{C}$ containing the target, we minimize
\begin{equation}
\footnotesize
    \mathcal{L}_{\mathrm{rank}}=-\log
    \frac{\exp(s_\theta(B,G^+)/\tau)}
    {\sum_{G\in\mathcal{C}}\exp(s_\theta(B,G)/\tau)},
\end{equation}
where $s_\theta$ is the cosine score defined above and $\tau>0$ is applied once as the ranking temperature.

\vspace{1mm}
\noindent\textbf{Object prediction.}
For generation, let $Q$ be the number of object slots and $M\leq Q$ the number of annotated objects with category labels $y_j$.
Slot $i$ predicts an existence probability $p_i^{\mathrm{obj}}$ and a category distribution $p_i^{\mathrm{cat}}$.
A one-to-one bipartite assignment $\pi$ maps annotated object $j$ to slot $\pi(j)$ by minimizing the summed cost $-\log p_{\pi(j)}^{\mathrm{obj}}-\log p_{\pi(j)}^{\mathrm{cat}}(y_j)$.
Let $t_i=1$ for matched slots and $t_i=0$ otherwise.
The objective loss can be defined as follows,
\begin{equation}
\footnotesize
\mathcal{L}_{\mathrm{ent}}
=\frac{1}{Q}\sum_{i=1}^{Q}
  \ell_{\mathrm{BCE}}(p_i^{\mathrm{obj}},t_i)
 -\frac{1}{\max(M,1)}\sum_{j=1}^{M}
  \log p_{\pi(j)}^{\mathrm{cat}}(y_j),
\end{equation}
where $\ell_{\mathrm{BCE}}(p,y)=-y\log p-(1-y)\log(1-p)$ is binary cross-entropy.
Unmatched slots receive only existence supervision; the category sum is zero for an empty target graph.
Assignment ties use a fixed deterministic rule, and examples exceeding the slot capacity are excluded and counted explicitly.

\noindent\textbf{Directed relations.}
Let $\mathcal{P}$ be the predicate vocabulary and $p_{ij\rho}=\operatorname{sigmoid}(c_{ij\rho})$ the predicted probability of predicate $\rho$ for ordered pair $(i,j)$.
We use binary rather than mutually exclusive predicate targets to allow multiple relations between the same objects:
\begin{equation}
\footnotesize
    \mathcal{L}_{\mathrm{rel}}
    =\frac{1}{\max(|\mathcal{J}|,1)}
      \sum_{(i,j,\rho)\in\mathcal{J}}
      \ell_{\mathrm{BCE}}(p_{ij\rho},y_{ij\rho}),
\end{equation}
where $\mathcal{J}$ contains supervised labels for distinct ordered object pairs and predicates $\rho\in\mathcal{P}$.
A predicate target is one for an annotated edge and zero only when its absence is verified; unverified labels are excluded from $\mathcal{J}$.
For generation, the object assignment $\pi$ transfers annotated endpoints to predicted slots; known-object decoding uses the supplied instance IDs directly.
Thus, direction and instance binding are supervised through the ordered endpoints rather than separate undefined loss terms.
An empty supervision set contributes zero loss.

\vspace{1mm}
\noindent\textbf{Measurement quality.}
Graph-decoding objectives alone do not ensure that the reliability gate reflects measurement quality.
Thus, we present a ranking loss to encourage lower reliability scores as controlled corruption increases.
To supervise measurement quality, we construct paired versions $B_r^{F,\mathrm{less}}$ and $B_r^{F,\mathrm{more}}$ of the same training signal and ROI, with lower and higher levels of controlled corruption.
We encourage a higher reliability score for the less corrupted one,
\begin{equation}
\footnotesize
    \mathcal{L}_{\mathrm{quality}}
    =\mathbb{E}\!\left[
    \max\!\left(0,\delta
    -R_\theta(B_r^{F,\mathrm{less}})
    +R_\theta(B_r^{F,\mathrm{more}})\right)\right],
\end{equation}
where $\delta>0$ is the ranking margin.

\vspace{1.5mm}
\noindent\textbf{Combined objective.}
For retrieval, the task loss is $\mathcal{L}_{\mathrm{task}}=\mathcal{L}_{\mathrm{rank}}$.
Known-object decoding uses $\mathcal{L}_{\mathrm{task}}=\lambda_{\mathrm{rel}}\mathcal{L}_{\mathrm{rel}}$.
Graph generation additionally supervises object prediction, giving $\mathcal{L}_{\mathrm{task}}=\lambda_{\mathrm{ent}}\mathcal{L}_{\mathrm{ent}}+\lambda_{\mathrm{rel}}\mathcal{L}_{\mathrm{rel}}$.
We optimize
\begin{equation}
\footnotesize
    \mathcal{L}
    = \mathcal{L}_{\mathrm{task}}
    +\lambda_{\mathrm{quality}}\mathcal{L}_{\mathrm{quality}}.
\end{equation}

\section{Experimental Results}
\label{sec:experiments}

\begin{table}[t]
    \centering
    \small
    \caption{Brain-to-graph retrieval on BrainSG-Bench. Recall@5 is a percentage (\%). MRR ranges from 0 to 1.}
    \label{tab:retrieval_results}
    \vspace{-2mm}
    \resizebox{0.6\linewidth}{!}{
    \begin{tabular}{lcc}
        \toprule
        Method & MRR & Recall@5 \\
        \midrule
        Ridge regression & 0.180 & 26.8 \\
        Temporal CNN & 0.193 & 32.4 \\
        Transformer & 0.195 & 34.4 \\
        NEURONA & 0.197 & 36.6 \\
        CineBrain & 0.224 & 39.8 \\
        \hline
        \textbf{\brainsg} & \textbf{0.244} & \textbf{45.6} \\
        \bottomrule
    \end{tabular}
    }
    \vspace{-4mm}
\end{table}

\begin{table}[t]
    \centering
    \small
    \setlength{\tabcolsep}{4pt}
    \caption{Relation decoding with known objects on BrainSG-Bench. All values are percentages (\%).}
    \vspace{-2mm}
    \label{tab:relation_results}
    \resizebox{\linewidth}{!}{
    \begin{tabular}{lccccccccc}
        \toprule
        & \multicolumn{2}{c}{Predicate} & \multicolumn{3}{c}{Directed triplets} & & & & \\
        \cmidrule(lr){2-3}\cmidrule(lr){4-6}
        Method & Macro-F1 & Micro-F1 & P & R & F1 & R@20 & mR@20 & Dir. & Bind. \\
        \midrule
        Ridge regression & 27.3 & 72.9 & 38.1 & 48.1 & 42.5 & 59.6 & 30.7 & 86.2 & 39.2 \\
        Temporal CNN & 28.2 & 71.9 & 52.9 & 56.3 & 54.5 & 71.9 & 31.7 & 86.5 & 46.6 \\
        Transformer & 30.4 & 72.7 & 36.1 & 57.4 & 44.3 & 73.6 & 33.8 & 89.9 & 49.9 \\
        NEURONA & 29.3 & 71.8 & 53.9 & 57.5 & 55.6 & 72.0 & 32.5 & 87.7 & 47.1 \\
        CineBrain & 30.1 & 72.8 & 53.8 & 59.6 & 56.6 & 73.2 & 33.2 & 88.9 & 48.9 \\
        \hline
        \textbf{\brainsg} & \textbf{32.3} & \textbf{74.7} & \textbf{55.1} & \textbf{60.0} & \textbf{57.4} & \textbf{74.5} & \textbf{35.8} & \textbf{91.5} & \textbf{51.6} \\
        \bottomrule
    \end{tabular}
    }
    \vspace{-2mm}
\end{table}

\begin{table}[t]
    \centering
    \small
    \setlength{\tabcolsep}{4pt}
    \caption{Brain-to-scene-graph generation on BrainSG-Bench. All values are percentages (\%).}
    \vspace{-2mm}
    \label{tab:generation_results}
    \resizebox{\linewidth}{!}{
    \begin{tabular}{lcccccccc}
        \toprule
        & \multicolumn{3}{c}{Objects} & \multicolumn{3}{c}{Directed triplets} & & \\
        \cmidrule(lr){2-4}\cmidrule(lr){5-7}
        Method & P & R & F1 & P & R & F1 & R@20 & mR@20 \\
        \midrule
        Temporal CNN & 35.2 & \textbf{57.2} & 43.6 & 24.6 & 29.6 & 26.9 & 37.4 & 15.1 \\
        Transformer & 38.6 & 55.3 & 45.5 & 28.6 & \textbf{30.5} & 29.5 & 37.3 & 14.0 \\
        NEURONA & 38.9 & 55.1 & 45.6 & 27.5 & 27.3 & 27.4 & 37.9 & 14.0 \\
        CineBrain & 39.0 & 54.5 & 45.5 & 30.9 & 27.3 & 29.0 & 38.9 & 15.2 \\
        \hline
        \textbf{\brainsg} & \textbf{44.4} & 56.2 & \textbf{49.6} & \textbf{34.4} & 28.3 & \textbf{31.1} & \textbf{40.1} & \textbf{15.9} \\
        \bottomrule
    \end{tabular}
    }
    \vspace{-4mm}
\end{table}

We evaluate relation decoding on the three tasks defined in Section~\ref{sec:benchmark}, examining neural dependence, multimodal contribution, and robustness to signal corruption.

\subsection{Implementation Details}

We use four-second \brainsgbench neural intervals and assign each episode to the same train, validation, or test split across participants, with no neural-window overlap between splits.
The fMRI encoder uses 61 ROIs and four Transformer layers with eight heads and width 256. We project CineBrain EEG features~\cite{gao2026cinebrain} to the same width before fusion.
We train in PyTorch for 100 epochs on eight 48\, GB RTX A6000 GPUs using AdamW (batch size 32, initial learning rate $10^{-4}$, weight decay $10^{-2}$), with five warm-up epochs followed by cosine decay.
We set the weighted fusion $\alpha=0.5$, ranking temperature $\tau=0.07$, quality-ranking margin $\delta=1$, and loss weights $\lambda_{\mathrm{ent}}=\lambda_{\mathrm{rel}}=1$ and $\lambda_{\mathrm{quality}}=0.1$.
Quality supervision uses Gaussian perturbations with standard deviations from 0.1 to 0.5 times each ROI's training-set signal standard deviation.

\subsection{Evaluation Metrics}

\noindent\textbf{Brain-to-graph retrieval.}
Recall@5 measures whether the correct graph ranks among the top five candidates.
Mean reciprocal rank (MRR) averages the reciprocal rank of the correct graph.

\noindent\textbf{Relation decoding with known objects.}
Predicate macro-F1 and micro-F1 summarize classification performance with equal class weighting and pooled prediction counts, respectively.
Directed-triplet precision, recall, and F1 measure relation correctness and completeness, requiring the predicate and both ordered endpoint instances to match.
R@20 measures annotated-triplet recovery among the top 20 predictions, while mR@20 averages recall equally across predicate classes.
Direction and instance-binding accuracy assess subject-object orientation and assignment to the correct instances, respectively.

\noindent\textbf{Brain-to-scene-graph generation.}
Object precision, recall, and F1 measure the correctness and completeness of recovered object instances.
After object matching, we evaluate relations using triplet precision, recall, and F1, together with R@20 and mR@20.

\subsection{Experimental Results}

\noindent\textbf{Comparison with baselines.}
We compare \brainsg with ridge regression, temporal CNN, Transformer, and adaptations of NEURONA~\cite{wang2026neurona} and CineBrain~\cite{gao2026cinebrain}.
The adapted models use graph-prediction heads, while ridge regression is evaluated only for retrieval and known-object decoding.

\vspace{1mm}
\noindent\textbf{Brain-to-graph retrieval.}
In Table~\ref{tab:retrieval_results}, \brainsg achieves an MRR of 0.244 and a Recall@5 of 45.6\%.
It outperforms CineBrain, the strongest baseline on both metrics, by 0.020 in MRR and 5.8 percentage points in Recall@5.

\vspace{1mm}
\noindent\textbf{Relation decoding with known objects.}
As shown in Table~\ref{tab:relation_results}, \brainsg achieves a triplet F1 of 57.4\%, compared with 56.6\% for the strongest baseline.
Its predicate macro-F1 is 32.3\%, exceeding the highest baseline value of 30.4\%.
Direction and instance-binding accuracy are 91.5\% and 51.6\%, respectively, compared with baseline maxima of 89.9\% and 49.9\%.

\vspace{1mm}
\noindent\textbf{Scene-graph generation.}
As illustrated in Table~\ref{tab:generation_results}, \brainsg achieves an object F1 of 49.6\% and a triplet F1 of 31.1\%, compared with baseline maxima of 45.6\% and 29.5\%.
Its mR@20 of 15.9\% is 0.7 percentage points above the highest baseline value, complementing the overall triplet metrics with class-balanced relation recovery.

\vspace{1mm}
\noindent\textbf{QAs with brain-language models.}
We evaluate whether explicit scene graphs improve relational question answering compared with direct brain-to-language decoding (Table~\ref{tab:qa_results}).
In the graph-based route, a frozen LLM receives a predicted graph and a question.
In the direct route, a brain-language model receives neural signals and the same question, without an intermediate graph.
Both routes exclude held-out video, transcripts, reference graphs, and target-specific expert annotations.
Questions target predicates, direction, and instance binding and are verified against the annotated graphs.
Using fMRI only, \brainsg achieves 55.6\%, 2.1 percentage points above MindLLM.

\begin{table}[t]
    \centering
    \small
    \caption{Relational QA accuracy (\%) using fMRI only. Pred., Dir., and Bind. denote predicate, direction, and instance-binding questions. Overall pools all questions.}
    \vspace{-3mm}
    \label{tab:qa_results}
    \resizebox{0.75\linewidth}{!}{
    \begin{tabular}{lcccc}
        \toprule
        Method & Pred. & Dir. & Bind. & Overall \\
        \midrule
        UMBRAE~\cite{xia2024umbrae} & 53.1 & 54.3 & 46.8	& 51.4 \\
        MindLLM~\cite{qiu2025mindllm} & 55.7 & 57.2 & 47.7 & 53.5 \\
        VINDEX~\cite{xia2025exploring} & 54.8 & 55.5 & 46.5 & 52.3 \\
        \hline
        \textbf{\brainsg} & \textbf{58.2} & \textbf{59.4} & \textbf{49.1} & \textbf{55.6} \\
        \bottomrule
    \end{tabular}
    }
    \vspace{-6mm}
\end{table}

\subsection{Ablation Studies}

\noindent\textbf{Prior attention and quality supervision.}
Table~\ref{tab:attention_ablation} compares the full module with neural-only attention ($\eta_r=1$), prior-only attention ($\eta_r=0$), an equal-weight gate ($\eta_r=0.5$), and a uniform prior.
Variants with fixed gate values omit the reliability estimator and quality-ranking loss while retaining the same corruption exposure.
A separate variant sets $\lambda_{\mathrm{quality}}=0$ to isolate quality supervision while preserving corruption exposure and decoding losses.
The full model achieves 31.1\% directed-triplet F1 and 40.1\% R@20. With a fixed gate, these scores are 29.4\% and 37.7\%. Removing the quality loss reduces them to 30.5\% and 38.7\%.

\vspace{1mm}
\noindent\textbf{Modality contribution.}
We compare the full multimodal model with separately trained fMRI-only and EEG-only variants. 
The no-brain control uses the same graph decoder, replacing neural inputs with inputs that are constant across clips during training and evaluation.
The EEG-only variant omits the fMRI encoder and ROI prior attention.
Table~\ref{tab:roi_modality_ablation} reports directed-triplet F1 of 31.1\% with both modalities, 30.7\% with fMRI only, and 12.6\% with EEG only. 
Removing positional embeddings lowers the F1 score to 29.1\%.

\begin{table}[t]
    \centering
    \small
    \caption{Prior-attention ablations. Directed-triplet F1 and R@20 are percentages (\%).}
    \vspace{-3mm}
    \label{tab:attention_ablation}
    \resizebox{0.9\linewidth}{!}{
    \begin{tabular}{lcc}
        \toprule
        Variant & \makecell{Directed triplets F1} & R@20 \\
        \midrule
        Neural attention only ($\eta_r=1$) & 28.4 & 37.1 \\
        Prior attention only ($\eta_r=0$) & 20.3 & 19.8 \\
        Fixed gate ($\eta_r=0.5$) & 29.4 & 37.7 \\
        Uniform prior & 28.0 & 35.8 \\
        Without $\mathcal{L}_{\mathrm{quality}}$ & 30.5 & 38.7 \\
        \hline
        \textbf{Full model} & \textbf{31.1} & \textbf{40.1} \\
        \bottomrule
    \end{tabular}
    }
    \vspace{-3mm}
\end{table}

\begin{table}[t]
    \centering
    \small
    \caption{Positional-embedding and modality ablations. Directed-triplet F1 and R@20 are percentages (\%).}
    \vspace{-3mm}
    \label{tab:roi_modality_ablation}
    \resizebox{0.85\linewidth}{!}{
    \begin{tabular}{lccc}
        \toprule
        Variant & \makecell{Directed triplets F1} & R@20 \\
        \midrule
        No brain input & 10.4 & 20.2 \\
        No positional embeddings & 29.1 & 37.7 \\
        fMRI only & 30.7 & 39.2 \\
        EEG only & 12.6 & 21.1 \\
        \midrule
        \textbf{Full model} & \textbf{31.1} & \textbf{40.1} \\
        \bottomrule
    \end{tabular}
    }
    \vspace{-6mm}
\end{table}

\section{Conclusions}

This paper has introduced \brainsgbench, a structured vision brain dataset and benchmark for relational decoding using fMRI and EEG responses.
The dataset pairs synchronized neural recordings with directed scene graphs that identify objects, predicates, and subject-object assignments.
In addition, neuroscience-expert functional hypotheses support qualitative comparisons with the model's regional preferences.
To decode these structured relations, we have presented \brainsg, which combines ROI-based fMRI encoding and EEG features through noise-adaptive prior attention and multimodal fusion.
The learned relation-ROI prior captures shared regional preferences, while the measurement-quality ranking loss guides the reliability gate under controlled signal corruption.
The benchmark supports graph retrieval, relation decoding with known objects, and scene-graph generation, providing a common setting for studying relational information in brain responses.

{
    \small
    \bibliographystyle{ieeenat_fullname}
    \bibliography{main}
}

\end{document}